\documentclass[lettersize,journal]{IEEEtran}
\usepackage{multicol}
\usepackage{epsfig}
\usepackage{color}
\usepackage{graphicx}
\usepackage{multirow}
\usepackage[table, x11names]{xcolor}
\usepackage{colortbl}
\usepackage{url}
\usepackage[rflt]{floatflt}
\usepackage{textcomp}
\usepackage[percent]{overpic}
\usepackage{wrapfig}
\usepackage{xspace}
\usepackage{booktabs}
\usepackage{tabularx}
\usepackage{adjustbox}
\usepackage{amsmath}
\usepackage[colorinlistoftodos,prependcaption,textsize=tiny,disable]{todonotes}
\usepackage{mathtools}
\usepackage{bm}
\usepackage{cuted}    
\usepackage{array}
\usepackage{lipsum}
\usepackage{amsfonts}
\usepackage{dsfont}
\usepackage{pgf}
\usepackage{adjustbox}

\makeatletter
\let\NAT@parse\undefined
\makeatother
\usepackage[numbers,sort&compress]{natbib}
\usepackage[
  breaklinks=true,
  colorlinks=true,
  linkcolor=blue,
  citecolor=blue,
  urlcolor=blue
]{hyperref}

\newcommand{\vf}[1]{{\bm{#1}}}
\newcommand{\mf}[1]{{\mathbf{#1}}}

\usepackage{xfp}
\usepackage{siunitx}

\makeatletter
\newcommand{\fullcellstrut}{%
  \rule[-\dp\@arstrutbox]{0pt}{\dimexpr\ht\@arstrutbox+\dp\@arstrutbox\relax}%
}
\makeatother

\DeclareRobustCommand{\srate}[3][0.8]{%
  \begingroup
  \pgfmathsetmacro{\p}{100*(#2)/(#3)}%
  \pgfmathsetmacro{\pp}{min(max(\p,0),100)}%
  \pgfmathsetmacro{\t}{\pp/100}%
  \pgfmathsetmacro{\amax}{#1}%
  \pgfmathparse{\t<0.5?1:0}%
  \ifnum\pgfmathresult=1\relax
    \pgfmathsetmacro{\w}{\amax*(2*\t*0.3 + 0.7)}
    \pgfmathtruncatemacro{\k}{round(100*\w)}%
    \edef\cellcol{white!\k!red}%
  \else
    \pgfmathsetmacro{\w}{\amax*((2 - 2*\t) * 0.5 + 0.7)}
    \pgfmathtruncatemacro{\k}{round(100*\w)}%
    \edef\cellcol{white!\k!blue}%
  \fi
  \expandafter\cellcolor\expandafter{\cellcol}%
  \fullcellstrut%
  \num[round-mode=places,round-precision=1]{\pp}%
  \endgroup
}

\makeatletter
\def\@IEEEauthorblockconfadjspace{-1.5em}
\makeatother

\usepackage{orcidlink}

\title{Right-Side-Out: Learning Zero-Shot Sim-to-Real Garment Reversal}

\IEEEaftertitletext{\vspace{-2.0em}}
\author{Chang Yu$^{*1}$\raisebox{0.3em}{\orcidlink{0009-0000-2613-9885}}, Siyu Ma$^{*1}$\raisebox{0.3em}{\orcidlink{0009-0002-7681-1670}}, Wenxin Du$^{1}$\raisebox{0.3em}{\orcidlink{0009-0006-1407-2007}}, Zeshun Zong$^{1}$\raisebox{0.3em}{\orcidlink{0000-0002-3256-1692}}, Han Xue$^{3}$\raisebox{0.3em}{\orcidlink{0000-0002-0847-1750}}, Wendi Chen$^{3}$\raisebox{0.3em}{\orcidlink{0009-0006-2128-4438}},
\\Cewu Lu$^{3}$\raisebox{0.3em}{\orcidlink{0000-0003-1533-8576}}, Yin Yang$^{4}$\raisebox{0.3em}{\orcidlink{0000-0001-7645-5931}}, Xuchen Han$^{2}$\raisebox{0.3em}{\orcidlink{0000-0003-4542-271X}}, Joseph Masterjohn$^{2}$\raisebox{0.3em}{\orcidlink{0000-0001-9605-7674}}, Alejandro Castro$^{2}$\raisebox{0.3em}{\orcidlink{0000-0003-3049-0066}}, Chenfanfu Jiang$^{1}$\raisebox{0.3em}{\orcidlink{0000-0003-3506-0583}}
\thanks{* equal contribution.}
\thanks{$^{1}${\tt\footnotesize g1n0st@live.com, siiyuma@outlook.com, \{wenxindu, zeshunzong, cffjiang\}@ucla.edu}, AIVC Lab, UCLA, USA.}
\thanks{$^{2}${\tt\footnotesize \{xuchen.han,joe.masterjohn,alejandro.castro\} @tri.global}, Toyota Research Institute, USA.}
\thanks{$^{3}${\tt\footnotesize \{xiaoxiaoxh, chenwendi-andy, lucewu\}@sjtu.edu.cn}, Shanghai Jiao Tong University, China.}
\thanks{$^{4}${\tt\footnotesize yin.yang@utah.edu}, University of Utah, USA.}
}

\begin{document}
\maketitle

\begin{abstract}
Turning garments right-side out is a challenging manipulation task: it is highly dynamic, entails rapid contact changes, and is subject to severe visual occlusion. We introduce \emph{Right-Side-Out}, a zero-shot sim-to-real framework that effectively solves this challenge by exploiting task structures. We decompose the task into \textsc{Drag}/\textsc{Fling} to create and stabilize an access opening, followed by \textsc{Insert\&Pull} to invert the garment. Each step uses a depth-inferred, keypoint-parameterized bimanual primitive that sharply reduces the action space while preserving robustness. Efficient data generation is enabled by our custom-built, high-fidelity, GPU-parallel Material Point Method (MPM) simulator that models thin-shell deformation and provides robust and efficient contact handling for batched rollouts. Built on the simulator, our fully automated pipeline scales data generation by randomizing garment geometry, material parameters, and viewpoints, producing depth, masks, and per-primitive keypoint labels without any human annotations. With a single depth camera, policies trained entirely in simulation deploy zero-shot on real hardware, achieving up to 81.3\% success rate. By employing task decomposition and high fidelity simulation, our framework enables tackling highly dynamic, severely occluded tasks without laborious human demonstrations. More details and supplementary material are on the website: \href{https://right-side-out.github.io}{https://right-side-out.github.io}.
\end{abstract}

\section{Introduction}

\IEEEPARstart{G}{arment} manipulation is a core challenge in robotics, complicated by cloth's high deformability, complex self-contact, and friction. While recent works have made significant progress in tasks such as folding~\cite{ganapathi2021learning, hietala2022learning}, smoothing~\cite{ha2022flingbot, canberk2022cloth}, and hanging~\cite{chen2025robohanger, chen2025graphgarment}, almost all approaches share a critical assumption: the garment is already right-side out. Nevertheless, flipping a garment to its proper orientation, a fundamental step in real-world garment workflows (e.g. laundry) and a prerequisite to all aforementioned tasks, is long overlooked. This task (coined \textit{Right-Side-Out}) restores the garment to its wearable state, exposes functional features, and reduces ambiguity for downstream tasks. Despite its ubiquity, this \textbf{dexterous, manually intensive} task remains largely unexplored in robotics.

\textit{Right-Side-Out} presents three main challenges. (1) \textbf{Mixed dynamics}: the naturally hybrid dynamics of the task -- alternating between quasi-static singulation and dynamic and momentum-assisted pull-throughs -- complicate the learning of a unified policy. (2) \textbf{Complex topology and contact}: the task requires manipulating the garment's topology by routing a gripper through a narrow, self-occluding boundary loop (e.g. the hem). It is also inherently bimanual: one arm anchors the clothes, while the other manipulates, imposing long-horizon gripper-cloth and arm-arm collision constraints. (3) \textbf{Garment variation}: high variance in garment type, size, and material results in a combinatorial explosion in configurations, complicating robust policy learning and making demonstration-based data collection prohibitively costly.

Recently, end-to-end learning-based paradigms have made rapid advances in garment manipulation. These approaches, from imitation learning~\cite{chi2023diffusion, zhao2023learning} to Vision-Language-Action (VLA) policies~\cite{black2410pi0}, have enabled end-to-end execution of complex tasks, such as shirt hanging in ALOHA Unleashed~\cite{zhao2024aloha}. However, they rely on massive teleoperation datasets, which are labor-intensive and expensive to collect.

\begin{figure}
    \centering
    \includegraphics[width=\linewidth]{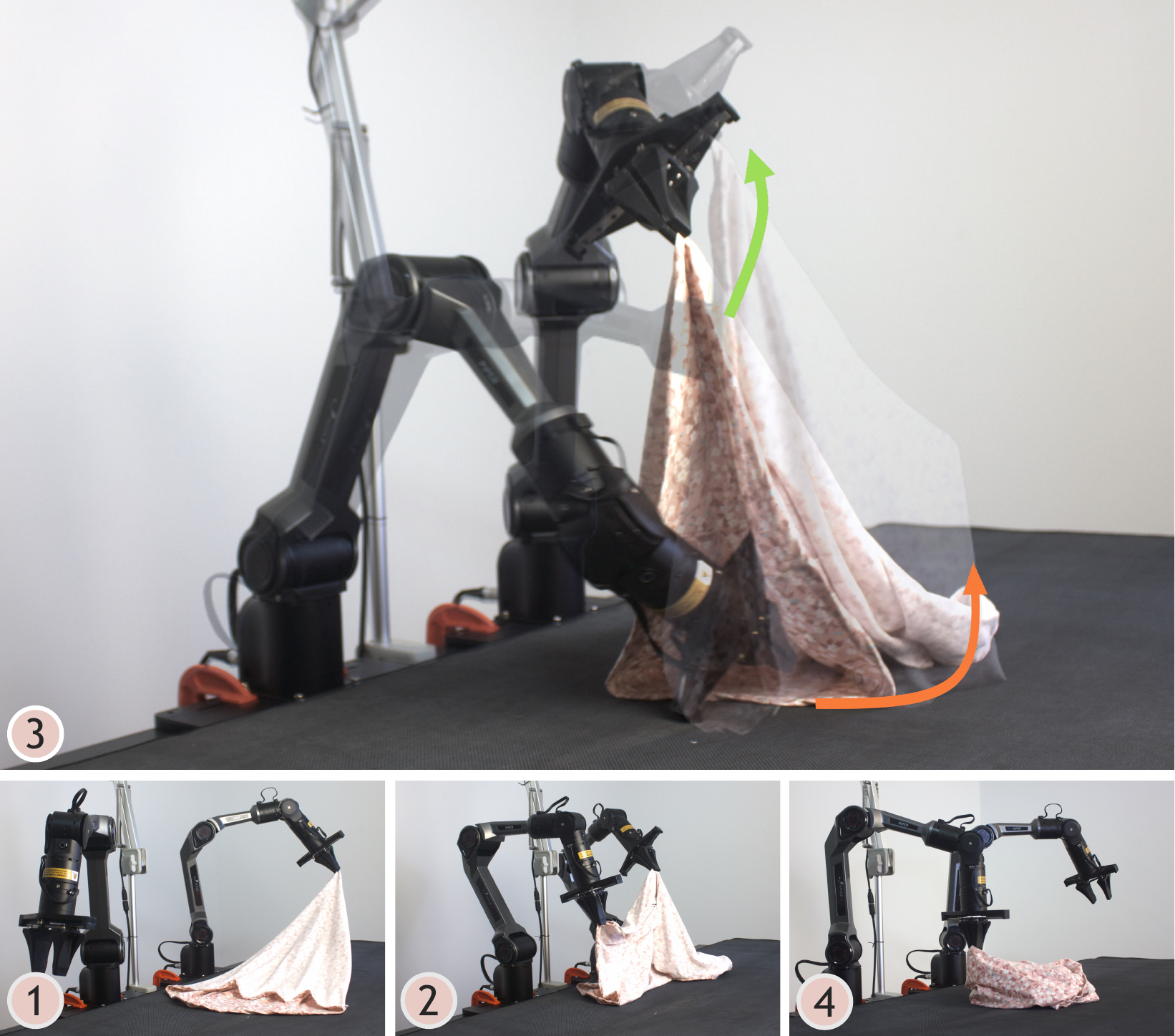}
\caption{Our system flips garments right-side-out with three keypoint-conditioned bimanual primitives: (1) \textsc{Drag} exposes a single-layer hem; (2) \textsc{Fling} brings the single layer to the top; (3) \textsc{Insert\&Pull} grasps the interior and pulls through; (4) final right-side-out state.}

    \label{fig:teaser}
\vspace{-5mm}
\end{figure}

Our key insight is to decompose this complex task into a sequence of simpler, human-inspired sub-goals: (i) layer singulation to create an access opening, and (ii) in-garment insertion and pull-through to complete the flip, as illustrated in Fig. \ref{fig:primitive}. Each step is implemented as a set of keypoint-conditioned primitives, predicted from depth and mask. This decomposition not only drastically reduces the action space compared to end-to-end methods, but also provides interpretable sub-goals and clear supervisions for robust learning.

In addition to the expensive labor cost of teleoperation, \textit{Right-Side-Out} also necessitates extensive sample coverage of critical contact events during intense interactions as manifested in Fig. \ref{fig:primitive}, which further complicates manual data collection. This motivates us to utilize faithful physical simulation for data-generation and policy learning. We developed a high-fidelity, Material Point Method (MPM)-based cloth simulator that accurately models frictional robot-cloth contact, self-collision, and codimensional deformations. Crucially, it supports batched multi-environment GPU simulation for scalable data generation. Our automated pipeline randomizes garment geometry, initial states, and material parameters to produce depth and mask labels without human annotation. Randomization in camera positions and parameters is also introduced for further data augmentation. 

Thanks to the high fidelity of our physics-based cloth simulator, the comprehensive coverage of our data augmentation strategy, and the strong robustness of our primitive-based model, we are able to achieve zero-shot sim-to-real transfer with \textit{only} one calibrated depth-camera: our policy is trained purely in simulation, and can be deployed on real robots without any additional fine-tuning. 

To summarize, our contributions are as follows:
\begin{itemize}
    \item We introduced the novel task of \textbf{robotic garment reversal}, referred to as  \textit{Right-Side-Out}.
    \item We proposed \textbf{parameterized bimanual action primitives} for layer singulation and in-garment pull-through, providing a compact yet robust action parameterization. 
    \item We developed a \textbf{high-fidelity, GPU-parallel MPM simulator} with a fully automatic data generation and annotation pipeline for cloth manipulation.
    \item We demonstrated \textbf{zero-shot sim-to-real transfer} on diverse garments.

\end{itemize}

\section{Related Work}

\subsection{Learning for Garment Manipulation}

Learning-based garment manipulation policies have attracted significant interest. Recent works span folding \cite{ganapathi2021learning, hietala2022learning, avigal2022speedfolding}, unfolding \cite{ha2022flingbot, canberk2022cloth}, hanging \cite{zhao2024aloha, chen2025robohanger, chen2025graphgarment}, and dressing~\cite{zhang2022learning, kotsovolis2024model}, often assuming canonical initial states~\cite{ha2022flingbot, lin2022learning}. Orthogonally, cloth singulation has been pursued with task-specific hardware~\cite{zhao2023flipbot, zhao2025learning}. Bimanual systems are increasingly employed to extend workspace and enable richer behavior, despite higher planning complexity and potential self-collisions. Beyond hardware considerations, action dynamics represent another key axis. Many pipelines remain quasi-static~\cite{ganapathi2021learning} for smoothing or folding, whereas unfolding from crumpled states has motivated dynamic actions (e.g., fling) to overcome occlusion and reach limits~\cite{ha2022flingbot}.

Garments are particularly difficult to manipulate due to their large configuration spaces, severe self-occlusions, and complex dynamics. Prior work has explored various learning paradigms to address these challenges. These include heuristic or keypoint-centric policies that detect features like wrinkles, corners, and edges \cite{sun2013heuristic, willimon2011model}; imitation learning from expert teleoperation \cite{black2410pi0, zhao2024aloha} and human preference \cite{xue2023unifolding, chen2024deformpam}; reinforcement learning, often bootstrapped by demonstrations or simplified goals \cite{matas2018sim, wu2019learning}; and goal-conditioned or self-supervised methods \cite{lee2021learning, seita2021learning}. In this work, we adopt depth-driven, keypoint-conditioned bimanual primitives that reduce the action space for the long-horizon \textit{Right-Side-Out} task, unifying quasi-static and dynamic operations within a single sim-to-real framework.

\begin{figure}
    \centering
    \includegraphics[width=\linewidth]{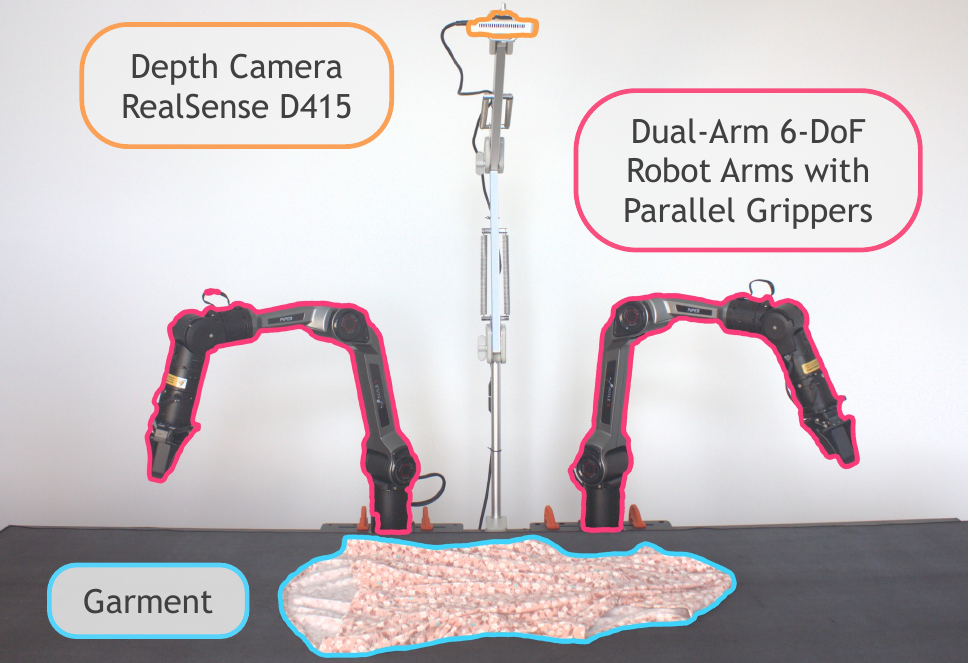}
    \caption{\textbf{Real-world robot setup.} Two 6-DoF arms and an overhead RealSense D415; see Section~\ref{sec:exp_setup}.}
    \vspace{-5mm}
    \label{fig:real_env_setting}
\end{figure}

\subsection{Physically-based Garment Dynamics}
Classical cloth solvers discretize garments as thin shells and must resolve \emph{codimensional} contact robustly. FEM pipelines typically rely on explicit collision handling with contact energies~\cite{bridson2002robust, harmon2008robust}. 
Incremental Potential Contact (IPC)~\cite{li2020incremental, Li2021CIPC, du2024embedded} introduces logarithmic contact barriers and continuous collision detection for intersection-free guarantees, at the cost of higher Newton iterations.
Position-based Dynamics (PBD)~\cite{muller2007position, macklin2016xpbd} trades accuracy for speed via constraint-based formulation and iterative solver, but slow convergence can accumulate unrecoverable penetrations in multi-layer contact. In contrast, the Material Point Method (MPM)~\cite{jiang2017anisotropic, han2019hybrid} uses a background Eulerian grid for particle–grid transfers, easing frictional contact and self-collision handling, and scales well on GPUs. Popular robotic simulation platforms reflect these trade-offs: PyFleX~\cite{macklin2014unified, li2018learning} and SoftGym~\cite{lin2021softgym} adopt PBD for real-time control benchmarks; IPC-based systems~\cite{du2024intersection, ma2025grip, li2025taccel} target intersection-free, two-way soft–rigid coupling; and MPM has been two-way coupled with rigid and integrated into Drake recently~\cite{zong2024convex, yu2025convex}.

Our work opts for a high-fidelity MPM simulator to provide scalable, simulation-only supervision -- without human demonstrations or annotation -- with physically faithful garment dynamics. To the best of our knowledge, we are the first to leverage an MPM-based solver to achieve zero-shot sim-to-real transfer in garment manipulation.

\section{Problem Statement}

\textbf{Settings.}
We consider a dual–arm robot with parallel grippers operating over a flat tabletop and observed by an overhead depth camera (Fig.~\ref{fig:real_env_setting}). The garment is distinguishable from the workspace, and the \emph{face polarity} (outside vs.\ inside) is visually identifiable from texture/appearance cues in the observation. We study sleeveless upper garments (e.g., tank tops) of varying size and shape; topologically, each is an orientable thin-shell surface whose boundary consists of four disjoint loops,
$\partial\Omega=\{\mathcal{C}_{\mathrm{hem}}, \mathcal{C}_{\mathrm{collar}}, \mathcal{C}_{\mathrm{arm}}^{L}, \mathcal{C}_{\mathrm{arm}}^{R}\}$, i.e., armholes are present but no attached sleeves. The garment is placed on the table \emph{inside-out} at the start of an episode.

\begin{figure}
    \centering
    \includegraphics[width=\linewidth]{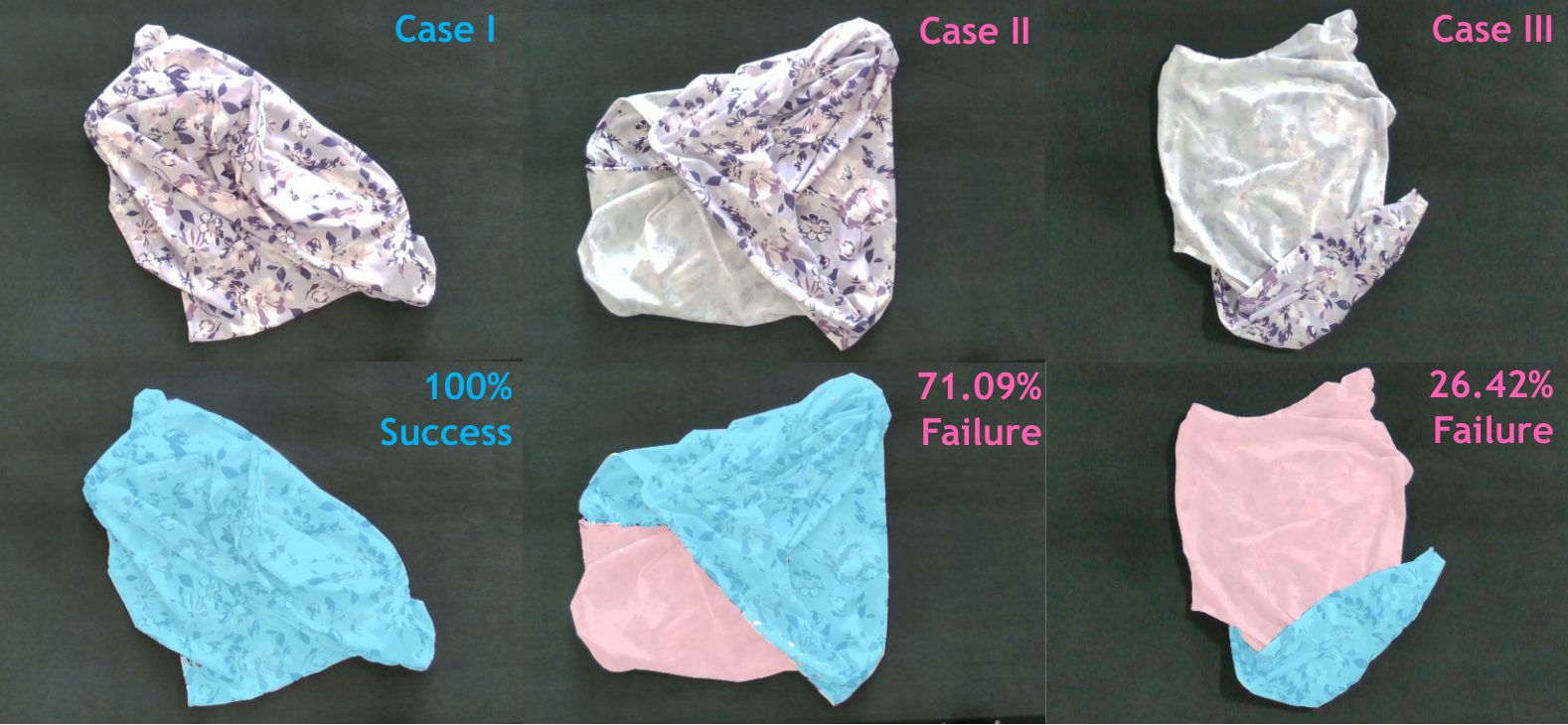}
    \caption{\textbf{Success metric.} From the top-down frame, we segment the garment with SAM~\cite{kirillov2023segment} and classify face polarity (outside, \textcolor{cyan}{blue}; inside, \textcolor{pink}{pink}). \emph{Right-side-out coverage} is the fraction of pixels within the garment mask labeled as outside; an episode is a success if coverage $\ge 0.80$. Case I succeeds (100\%); Cases II and~III fail (71.09\%, 26.42\%).}
    \vspace{-5mm}
    \label{fig:criteria}
\end{figure}

\textbf{Task and success metric.}
The objective is to control the bimanual robot to produce a final configuration in which the garment is right-side out. We evaluate performance using a binary success metric computed from the final top-down frame. The metric is based on \emph{right-side-out coverage}, defined as the fraction of pixels within the garment's mask whose visible face is classified as the outside surface.
An episode is considered a success if the coverage rate is $\ge$ 80\%; otherwise, it is a failure, as illustrated in Fig.~\ref{fig:criteria}.

To ensure our metric focuses on the flipping task alone, we initialize garments in a canonicalized pose (following~\cite{canberk2022cloth}) but with an inverted (inside-out) face polarity. After the policy runs, we allow an optional canonicalization step that preserves face polarity (e.g., light shaking or flinging~\cite{ha2022flingbot}) before measurement. This ensures the garment is in a clean state for downstream tasks.

\begin{figure*}
\centering
\includegraphics[width=\linewidth]{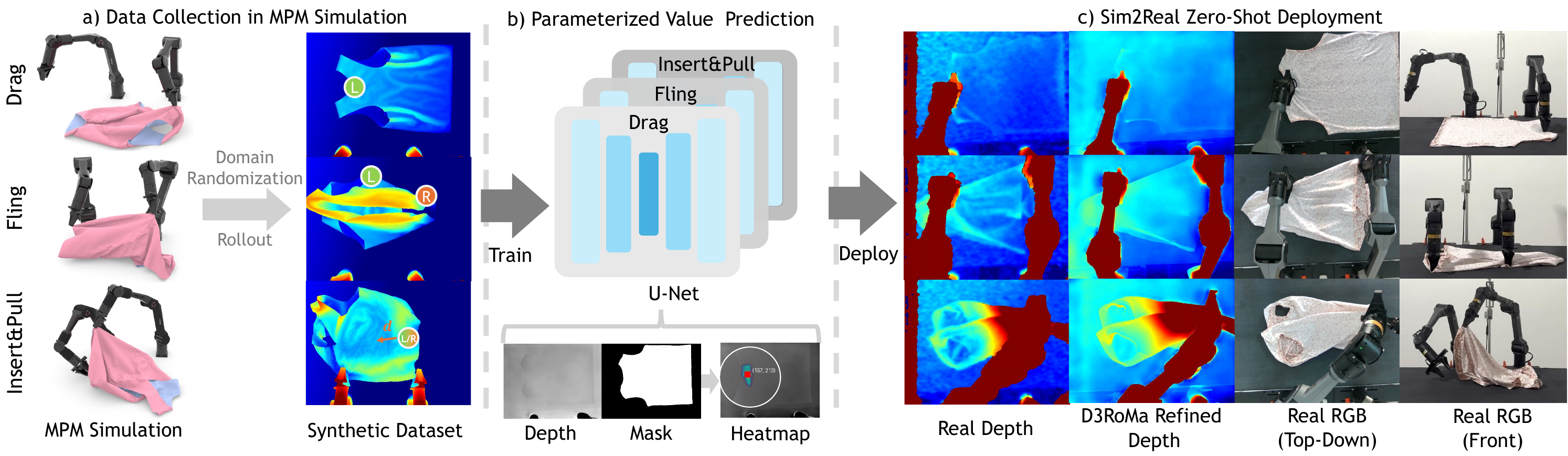}
\caption{\textbf{System overview:} (a) MPM simulation executes bimanual primitives on domain-randomized garments to generate synthetic depth, masks, and supervision; (b) per-primitive U-Nets take depth and mask as input and predict dense value maps for keypoint selection; (c) the same networks deploy zero-shot on real data using D3RoMa~\cite{wei2024d} and SAM masks~\cite{kirillov2023segment}.}

\vspace{-5mm}
\label{fig:pipeline}
\end{figure*}

\section{Method}

We outline our framework as follows. Section~\ref{sec:mpm} describes our multi-environment GPU-parallel MPM simulator with efficient collision handling and physically faithful garment dynamics. Section~\ref{sec:parameterization} defines the policy decomposition, where we decompose \emph{Right-Side-Out} into \textsc{Drag}, \textsc{Fling} and \textsc{Insert\&Pull}. Section~\ref{sec:unet} presents the action-primitive networks that take depth and mask as input and predict per-primitive value maps. Finally, Section~\ref{sec:sim2real} details the sim-to-real transfer for zero-shot deployment. The overall pipeline is illustrated in Fig.~\ref{fig:pipeline}.

\subsection{MPM Garment Simulation Environment}
\label{sec:mpm}

We introduce a GPU–parallel Material Point Method (MPM) environment tailored for thin-shell garments and robotic manipulations. Our codimensional MPM solver offers robust collision handling and excellent parallel efficiency: its regular grid and localized memory access avoid irregular contact traversals, yielding good load balance and scaling to multi-environments. We use the anisotropic elastoplastic constitutive model for codimensional materials (cloth) so that frictional contact of garment is expressed directly in the continuum; in practice, contact processing happens on the grid and requires no explicit collision queries.

\paragraph{State and discretization}
Let $\mathcal{S}$ be a triangulated garment surface with $N_v$ vertices and $N_f$ faces. The stacked vertices position vector is 
$\mf{x}_p=\big[\vf{x}_{p, 1}^\top,\dots,\vf{x}_{p, N_v}^\top\big]^\top\!\in\mathbb{R}^{3N_v}$.
For each of the $N_f$ triangular faces, one quadrature point is chosen at its centroid, yielding $N_f$ quadrature particles $\mf{x}_c=\big[\vf{x}_{c,1}^\top,\dots,\vf{x}_{c, N_f}^\top\big]^\top\!\in\mathbb{R}^{3N_f}$. They are evolved with a background Eulerian grid of spacing $h>0$ via APIC transfers~\cite{jiang2015affine}. Particles carry mass, elastic deformation $\mf{F}$, and material parameters; the grid nodes $\{i\}$ hold the DOFs velocities $\vf{v}_i$ on which boundary conditions are imposed. The in-plane eneregy for the garment is modeled as hyperelastic. We resolve frictional contact for self-collision and inter-garment interactions using the anisotropic elastoplastic cloth model of \cite{jiang2017anisotropic}. We refer readers to \cite{jiang2015affine} for MPM details. 

\paragraph{Garment-garment contact modeling}
Garment friction is formulated as a Coulomb-type yield condition on the Cauchy stress $\boldsymbol{\sigma}$ in the continuum \cite{jiang2017anisotropic}. Let $\vf{n}$ denote the unit normal vector to the local garment manifold. We define a codimensional Coulomb yield function
\begin{equation}
\mathcal{Y}(\boldsymbol{\sigma};\vf{n})
:= \big\|\big(\mf{I}-\vf{n}\vf{n}^\top\big)\boldsymbol{\sigma}\vf{n}\big\|
- \mu_{\mathrm{cloth}}\,\vf{n}^\top\boldsymbol{\sigma}\vf{n}
\;\le\; 0,
\label{eq:self_friction_yield}
\end{equation}

which states that the magnitude of tangential traction is bounded by the friction coefficient $\mu_{\mathrm{cloth}}$ times the compressive normal traction. At each step, we compute a trial elastic state and apply a return-mapping projection with no change to in-manifold strain, but rescales the shear components orthogonal to the surface so that Eq. \eqref{eq:self_friction_yield} holds. This yields sticking, sliding, and separation behavior, consistent with Coulomb friction for codimensional garment. 

\paragraph{Garment–robot contact modeling}

We treat each robot link $\ell$ as well as the workzone table as an immovable collider and enforce contact by a projected Coulomb update on grid velocities by one-way coupling.
At a contacted grid node $i$ with $\phi_\ell(\vf{x}_i)\!\le\!0$ and normal $\vf{n}$, let
$\vf{u}=\vf{v}_i-\vf{v}_\ell(\vf{x}_i)$, $u_n=\vf{u}\!\cdot\!\vf{n}$, and $\vf{u}_t=\vf{u}-u_n\vf{n}$.
The post-contact grid velocity $\vf{v}_i^{+}$ is then
\[
\vf{v}_i^{+} =
\begin{cases}
\displaystyle
\begin{aligned}
\vf{v}_i - u_n\,\vf{n}
-\min\!\Big(\|\vf{u}_t\|,\;\mu_{\mathrm{robot}}(-u_n)\Big)\,
\dfrac{\vf{u}_t}{\|\vf{u}_t\|+\varepsilon}, \\
\text{if } \phi_\ell(\vf{x}_i)\le 0~\text{and } u_n<0,
\end{aligned} \\
\vf{v}_i, \quad\quad\quad\quad\quad\quad\quad\quad\quad\quad\quad\quad\quad\quad~~~\text{otherwise,}
\end{cases}
\]
where $\mu_{\mathrm{robot}}$ is the friction coefficient and regularization $\varepsilon>0$. The above design yields a contact-robust, GPU-efficient simulator for large-scale rollout of long-horizon garment interactions with realistic thin-shell deformations. We evaluate speed and fidelity in Section \ref{sec:sim_eval}.

\subsection{Policy Parameterization}
\label{sec:parameterization}
\begin{figure*}
\centering
\includegraphics[width=\linewidth]{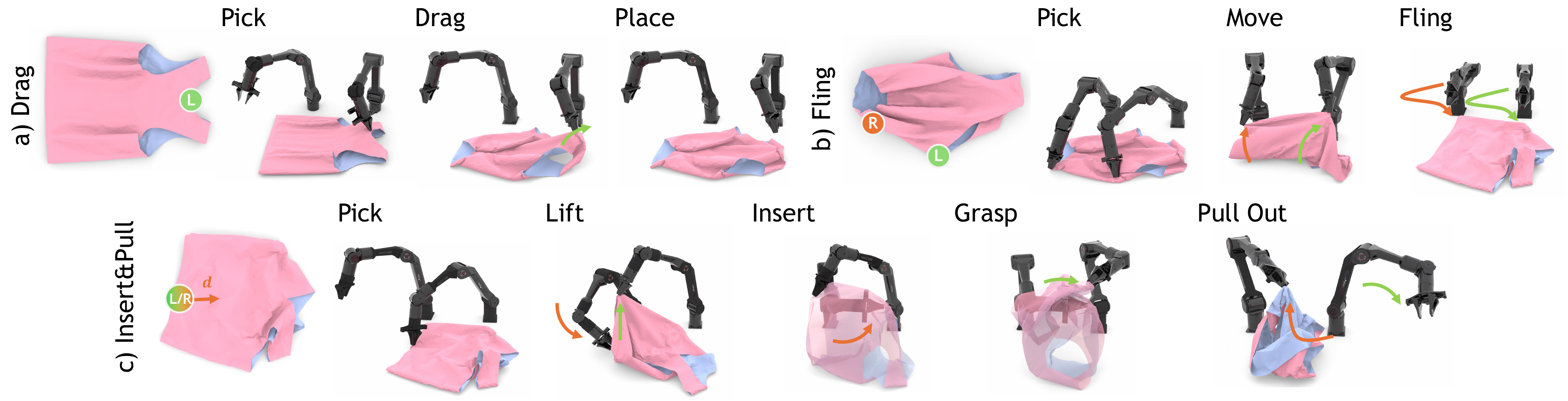}
\caption{\textbf{Primitives parameterization.} (a) \textsc{Drag}, (b) \textsc{Fling}, (c) \textsc{Insert\&Pull}; see Section~\ref{sec:parameterization}.}
\vspace{-5mm}
\label{fig:primitive}
\end{figure*}

We reduce the long-horizon, contact-rich problem to parameterized bimanual primitives instantiated by a few 2D image-plane keypoints from a calibrated top-down depth view captured by an overhead camera. This simple yet effective approach drastically reduces the action space, while also facilitates sim-to-real transfer. Each primitive $m\!\in\!\{\textsc{Drag},\textsc{Fling},\textsc{Insert\&Pull}\}$ is described by a few 2D keypoints, which are predicted from the depth images and masks. Those 2D keypoints are then mapped back to world space, to trigger end-effector motions with fixed templates. We detail each primitive action below. See also Fig.~\ref{fig:primitive} for a step-by-step illustration.

The first step, denoted the singulation stage, is to form a drag point near the hem where the front and back layers are separated. This allows one robot arm to lift the front layer, creating an access opening for the insertion of the second arm. Most prior singulation methods rely on specialized hardware. \cite{chen2023bagging} recognizes single layers with parallel grippers without extra sensors, but its bag-specific perception and grasp-height sensitivity hinder transfer to diverse garments. We observe that, in a canonical pose, the front neckline typically lies below the back neckline because of the front neck drop for anatomical fit. Although the collar loop $\mathcal{C}_{\mathrm{collar}}$ is typically too small for insertion, we can ``transport'' this single-layer patch towards the hem to create a larger opening, using a drag followed by a brief fling. 

For each pixel $u$ in the image plane, let $L(u)$ denote the number of garment layers at pixel $u.$ For instance, around the collar where the back neckline is exposed, $L(u) = 1$, whereas regions around the chest typically have $L(u)=2.$

\textbf{\textsc{Drag}.}
From a pick point $p_{\mathrm{drag}}$ in the single layer region of the collar $\{u\in \text{collar}:L(u)=1\}$, the robot drags the garment a fixed distance upward and then outward, using workspace friction to separate the two layers and advect the single-layer patch towards the hem region. That is, after this step, $\exists\,u \in \text{hem} \;\text{s.t.}\; L(u)=1$.

\textbf{\textsc{Fling}.}
To successfully create an opening with one arm, the hem must have a single-layer on the upper side, i.e. there exists a neighborhood $N$ of pixels s.t. $\forall u \in N, L(u) = 1$ and this layer is the upper layer (i.e. the front layer). We achieve this by grasping a hem-spanning pair $(p_L,p_R)$ and executing a fixed stretch-and-fling: lift to a set height, stretch to a force threshold, then apply a brief forward–backward pulse. Following prior dynamic-cloth work \cite{ha2022flingbot}, we fix the fling kinematics and only learn the grasp pair. This action is parameterized by $(p_L,p_R)$.

\textbf{\textsc{Insert\&Pull}. }
This action is parameterized by $(p_\mathrm{lift}, p_\mathrm{pull})$. 
The left arm grasps a point $p_\mathrm{lift} \in N$ and lifts a fixed distance upwards to $p'_\mathrm{lift},$ to create an access opening for subsequent insertion. The right arm positions its gripper directly beneath the lifted $p'_\mathrm{lift}$ (where the opening is largest), keeping its jaws parallel to the workspace, slides along $\vf{d}=p_\mathrm{pull}-p_\mathrm{lift}$ to $p_\mathrm{pull}$, raises for a fixed distance and closes the jaw. Gravity drapes the garment into the jaw, yielding a stable grasp; a fixed kinematic pull-out along $\vf{d}$ completes extraction.

\subsection{Action Primitive Networks}
\label{sec:unet}

We learn dense per-pixel action-value maps with a U-Net backbone \cite{ronneberger2015u}. Given an overhead observation $O\in\mathbb{R}^{C\times H\times W}$ (depth + binary masks) at the start of each primitive, the network outputs an $H\times W$ map whose argmax yields a pixel-wise target. Pixels are back-projected to 3D; controllers fix the $z$ height and kinematic profile, so networks predict only planar $(x,y)$.

\textbf{Unimanual primitive (\textsc{Drag}).}
Using collar annotations and a ray intersector, we deterministically form a dense supervision map
$Y[u]=\mathds{1}\{u\in \text{collar}: L(u)=1\}$.
We train a single network $Q_{\mathrm{drag}}$ with pixel-wise binary cross-entropy loss (BCE):
\(
\mathcal{L}_{\mathrm{drag}}=\mathrm{BCE}\big(Q_{\mathrm{drag}}[\cdot\,;O],\,Y\big).
\)

\textbf{Bimanual primitives (\textsc{Fling}, \textsc{Insert\&Pull}).} For bimanual primitives specified by a pair $(p_a,p_b),$ we use two networks sequentially: $Q_a(O)$ to select $p_a$, and $Q_b(O,p_a)$ to select $p_b$ conditioned on $p_a,$ via concatenating a one-hot map of $p_a$ to $O$. We optimize \[ \mathcal{L}_b = \mathrm{BCE}\big(Q_b[p_b;O,p_a],\,S\big), \mathcal{L}_a = \mathrm{BCE}\big(Q_a[p_a;O],\,\hat S\big) \] with $\hat S = \mathds{1}\!\left\{\max_{y}Q_b[y;O,p_a]>\tau\right\}$ and $\tau{=}0.5$. We heuristically sample candidates within annotated garment regions. 
Stage success $S$ is evaluated in simulation for each rollout. We instantiate: \begin{itemize} 
\item \textsc{Fling:} $(p_a,p_b)=(p_L,p_R)$, $(Q_a,Q_b)=(Q_{\mathrm{fling}}^L,Q_{\mathrm{fling}}^R)$; $S{=}1$ iff the area of the hem single-layer set $\{u:L(u)=1\}$ exceeds a threshold $\,\tau_{\mathrm{area}}$.
\item \textsc{Insert\&Pull:} $(p_a,p_b)=(p_{\mathrm{lift}},p_{\mathrm{pull}})$, $(Q_a,Q_b)=(Q_{\mathrm{lift}},Q_{\mathrm{pull}})$; $S{=}1$ iff \textit{right-side-out coverage} $\ge 0.80$ on mask (illustrated in Fig.~\ref{fig:criteria}).
\end{itemize}

\subsection{Sim-to-Real Transfer and Deployment}
\label{sec:sim2real}

To mitigate the physics gap, we apply domain randomization to material and contact parameters: Young’s modulus $E\in[300,3000]$\,kPa, density $\rho\in[1000,3000]$\,kg/m$^3$, friction coefficients $\{\mu_{\mathrm{cloth}}, \mu_{\mathrm{ground}}, \mu_{\mathrm{robot}}\}\in[0.3, 0.95]$, and RPIC damping $c\in[0.0,0.5]$; shearing stiffness $\gamma$ is sampled as a fixed fraction of $E$. For geometric variation, we procedurally generate garments with diverse $\mathcal{C}_{\mathrm{hem}}$ and $\mathcal{C}_{\mathrm{collar}}$ shapes, assign rule-based component annotations, and randomize overall size. To reduce the visual gap, we randomize camera intrinsics/extrinsics and add noise to depth and segmentation masks during training. At deployment, we use a well-calibrated overhead camera and D3RoMa-refined depth maps~\cite{wei2024d} to suppress depth noise. Without any real-world fine-tuning, the simulation-trained policy can be deployed on a dual-arm robot, achieving zero-shot sim-to-real transfer.

\begin{figure} [b]
    \centering
    \vspace{-5mm}
    \includegraphics[width=\linewidth]{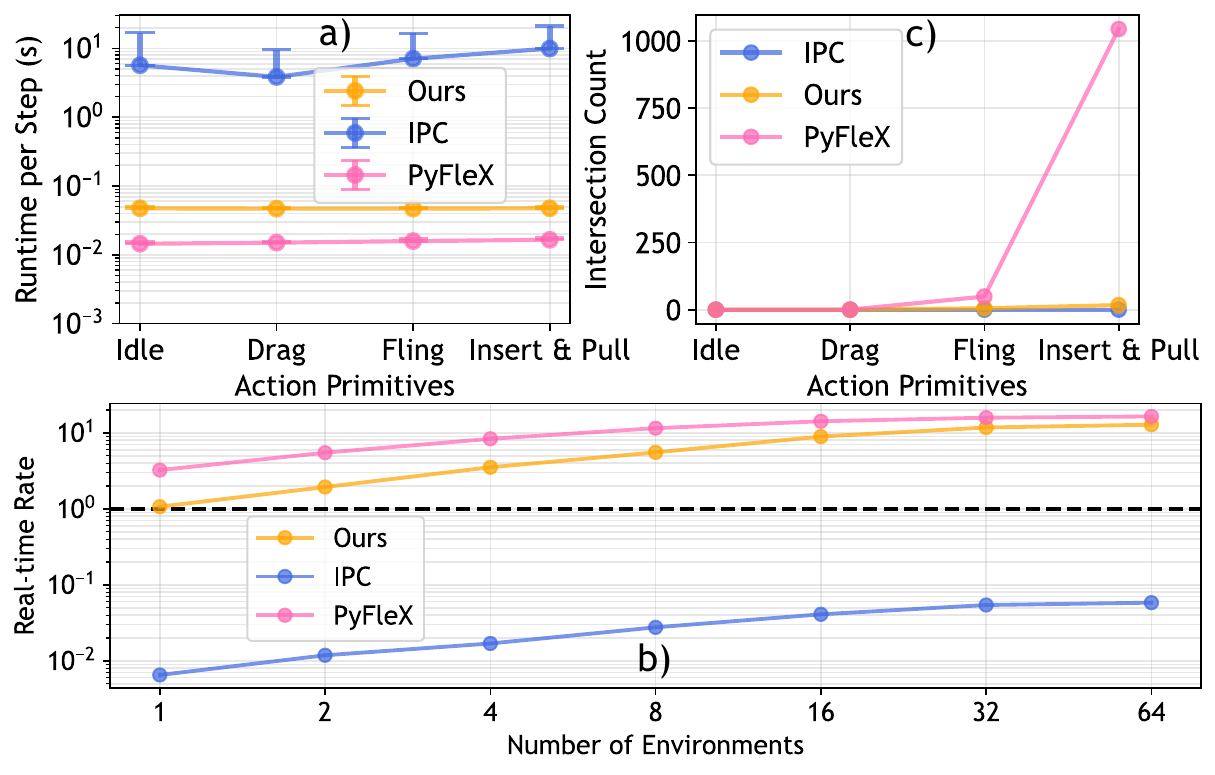}
    \caption{\textbf{Profiling.} (a) per-step runtime by primitive; (b) real-time rate vs.\ parallel environments (dashed = $1\times$); (c) max triangle--triangle intersections per step.}\label{fig:profiling}
\end{figure}

\begin{figure*}
\centering
\includegraphics[width=\linewidth]{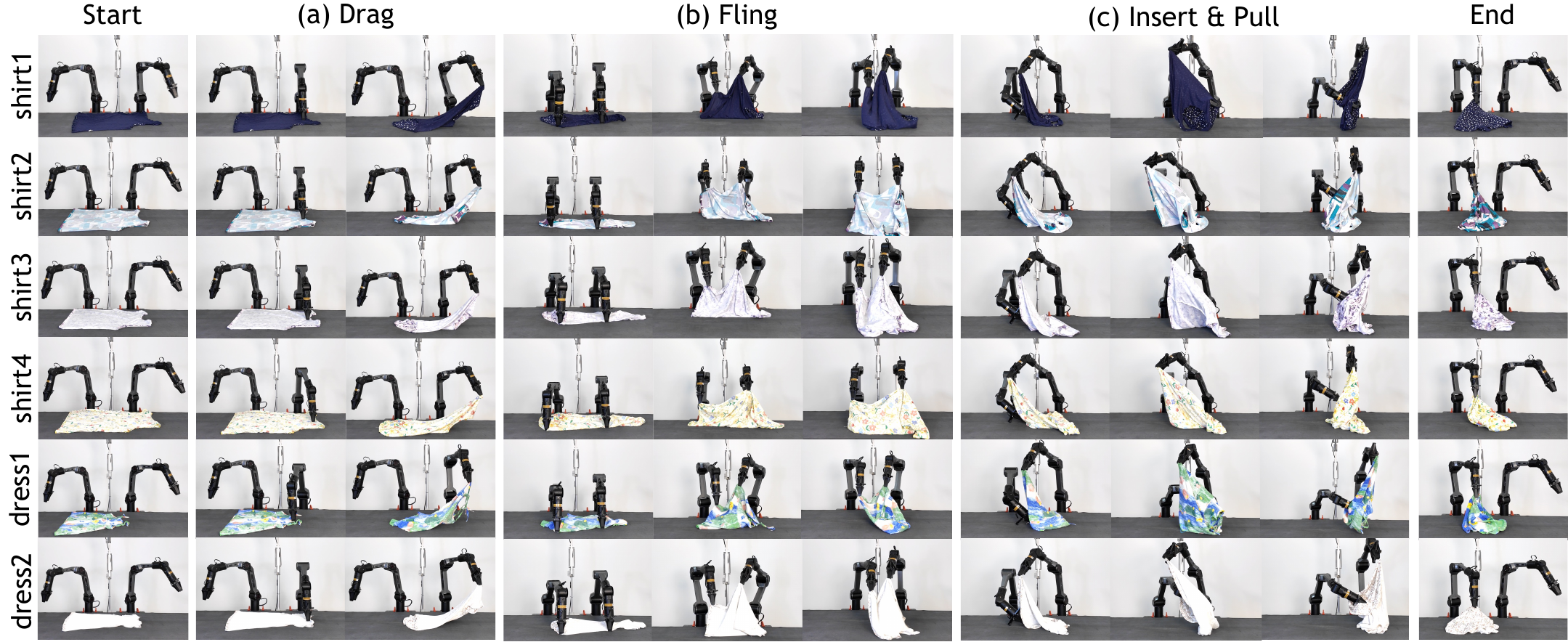}
\caption{\textbf{Results.} Real-world deployment of our system across diverse garments.}
\vspace{-5mm}
\label{fig:results}
\end{figure*}

\section{Experiments}

\subsection{Experimental Setup}
\label{sec:exp_setup}

\paragraph{Real-world deployment}
Our real-world experiment setup consists of two 6-DoF AgileX PiPER arms with parallel grippers. The top-down RGB-D image is captured with an Intel RealSense D415, generating observations \(O \in \mathbb{R}^{640 \times 480}\). A 3cm foam mat is placed on the workspace to allow the grippers to reach below the garment without colliding, as shown in Fig.~\ref{fig:real_env_setting}. Action primitives are executed at a control frequency of 20 Hz.

\paragraph{Simulation data collection} 

All simulation and training are run on a single NVIDIA RTX 4090 (24 GB). The control frequency is aligned with the real-world setup and the simulation takes a timestep $\Delta t=7\times10^{-4}$s with explicit integration under the CFL restriction. Using 64 parallel environments, a heuristic policy collects 20{,}000 trajectories with 128 synthetic garment meshes; end-to-end data generation takes approximately 25 GPU-hours. We render aligned depth images and garment masks with the SAPIEN renderer~\cite{xiang2020sapien}. For each primitive, we train a standard U-Net with 5 levels of downsampling/upsampling for 400 epochs; a full training run for each network takes about 1 hour.

\subsection{Simulator Performance and Fidelity Evaluation}
\label{sec:sim_eval}

We evaluate the throughput and physical fidelity of our MPM simulator against two commonly used baselines for deformable manipulation: (i) a multi–environment IPC solver~\cite{ma2025grip, li2025taccel} and (ii) PyFleX~\cite{macklin2014unified, li2018learning}, adopted in prior Sim-to-Real garment pipelines~\cite{chen2025robohanger, ha2022flingbot}. For a fair comparison, we only change the physics backend while keeping the data-generation policy, controllers, and control rate fixed. We run IPC at a timestep equal to the control step, owing to its Newton-based solver and continuous collision handling. PyFleX uses its default time step size $\Delta t=0.01\,$s.

We benchmark three aspects: (1) single-environment runtime per physics step at a 20\,Hz control loop; (2) real-time rate as the number of parallel environments increases; and (3) garment self-penetration, measured as the maximum number of triangle–triangle intersections per step during the rollout. Results are summarized in Fig.~\ref{fig:profiling}.

Our MPM simulator achieves per-step runtime comparable to PyFleX across primitives, below 50\,ms per step. IPC is about $10^2\times$ slower, making large-scale data generation impractical for long-horizon interaction. Real-time rate grows linearly with the number of environments for our simulator, reaching roughly 10$\times$ real-time rate at 64 environments. IPC remains sub-real-time rate even with many environments, reflecting its heavier contact solve.
Fidelity difference is most pronounced in the contact-rich \textsc{Insert\&Pull} stage. IPC, by construction, enforces non-penetration, though at a substantial computational cost. PyFleX exhibits severe interpenetration ($>10^3$), whereas our MPM simulator has much fewer penetrations ($<20$) in \textsc{Insert\&Pull}, sufficient to execute the primitive robustly.

Overall, our GPU-parallel MPM simulator provides a balanced operating point, combining fast rollouts with reliable handling of frictional and self-collision-rich interactions, essential for \emph{Right-Side-Out}.

\begin{table}[b]
\centering
\vspace{-5mm}
\caption{Method comparison in simulation.}
\setlength{\aboverulesep}{0pt}
\setlength{\belowrulesep}{0pt}
\setlength{\abovetopsep}{0pt}
\setlength{\belowbottomsep}{0pt}
\setlength{\cmidrulesep}{0pt}
\begin{tabular}{r|c|cccc}
\toprule
\textbf{Methods} & \textbf{Output} & $S_1$(\%)~$\uparrow$ & $S_2$(\%)~$\uparrow$ & $S_3$(\%)~$\uparrow$ & $S$(\%)~$\uparrow$ \\
\midrule
Heuristic     & Prim  & {\srate{16}{16}} & {\srate{13}{16}} & {\srate{8}{13}} & {\srate{8}{16}} \\
Ours          & Prim  & {\srate{16}{16}} & {\srate{15}{16}}  & {\srate{13}{15}} & {\srate{13}{16}} \\
DP~\cite{chi2023diffusion}       & Joint & {\srate{14}{16}} & {\srate{7}{14}} & {\srate{3}{7}} & {\srate{3}{16}} \\
ACT~\cite{zhao2023learning}      & Joint & {\srate{15}{16}} & {\srate{10}{15}} & {\srate{5}{9}} & {\srate{5}{16}} \\
\bottomrule
\end{tabular}
\label{tab:compare_sim}
\end{table}

\subsection{Policy Comparison in Simulation}
\label{sec:policy_eval_sim}

For each garment we run 16 trials and report the final success rate $S$ together with three stage-conditional rates $S_1$, $S_2$, $S_3$. $S$ counts episodes whose final state achieves right-side-out coverage (Fig.~\ref{fig:criteria}). $S_1$ is the success rate of \textsc{Drag}, producing a hem-side single-layer region; $S_2$ is the success rate of \textsc{Fling}, exposing an upper-side hem aperture, conditioned on $S_1$ success; and $S_3$ is the final success rate conditioned on $S_1$ and $S_2$. Note that $S=S_1 \times S_2 \times S_3$.

\begin{figure}[t]
    \centering
    \includegraphics[width=\linewidth]{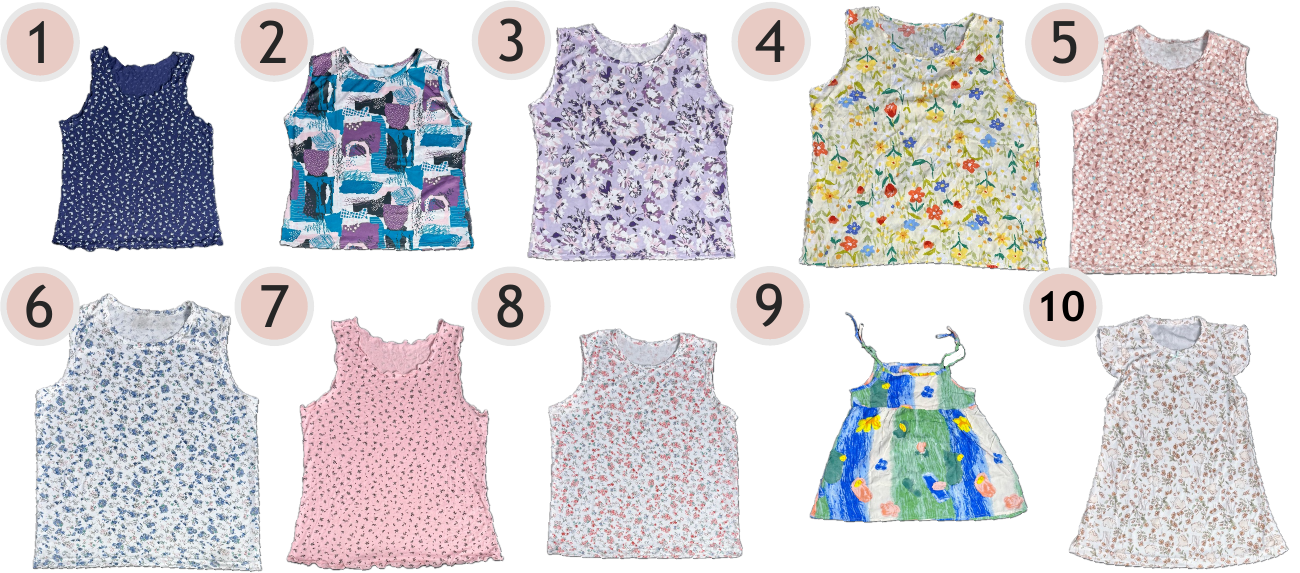}
    \caption{\textbf{Real-world garment assets.} The garments in experiment vary in type, size, color, fabric, and collar/hem shapes.}
    \vspace{-5mm}
    \label{fig:clothes_headshots}
\end{figure}

We evaluate the policy on unseen garments in simulation; the results are reported in Table~\ref{tab:compare_sim}. The \emph{Heuristic} baseline is a rule-based controller that utilizes the garment mesh state; these manually specified rules underperform our learned primitive policies, which were trained on a large amount of data with supervised labels, collected and evaluated using our simulator. We also train two end-to-end imitation learning methods, Action-chunked Transformer (ACT)~\cite{zhao2023learning} and Diffusion Policy (DP)~\cite{chi2023diffusion}, using the successful portion of the same data with the same perception. These benchmarks take the per-frame observation as input and output 14-DoF bimanual joint actions at 20Hz, producing around $800$ action steps per episode. In contrast, our method predicts only five image-plane keypoints to parameterize the primitives. While ACT and DP learn \textsc{Drag} and \textsc{Fling} reasonably well, they struggle in the long-horizon, contact-rich \textsc{Insert\&Pull} stage under severe occlusions and diverse observation states.

\subsection{Policy Comparison in Real-world}
\label{sec:policy_eval_real}

We evaluate the zero-shot transferability of the methods in Section~\ref{sec:policy_eval_sim} by training only in simulation and deploying on hardware without any fine-tuning. We test four garments (\#1–\#4 in Fig.~\ref{fig:clothes_headshots}) under identical initializations and action budgets for a fair comparison. Results are in Table~\ref{tab:compare_real}.

\begin{table}[b]
\vspace{-5mm}
\centering
\caption{Method comparison in real-world.}
\setlength{\aboverulesep}{0pt}
\setlength{\belowrulesep}{0pt}
\setlength{\abovetopsep}{0pt}
\setlength{\belowbottomsep}{0pt}
\setlength{\cmidrulesep}{0pt}
\begin{tabular}{r|cccc}
\toprule
\textbf{Methods} & $S_1$(\%)~$\uparrow$ & $S_2$(\%)~$\uparrow$ & $S_3$(\%)~$\uparrow$ & $S$(\%)~$\uparrow$ \\
\midrule
ACT~\cite{zhao2023learning}      & {\srate{2}{16}} & {\srate{0}{2}} & {\srate{0}{16}} & {\srate{0}{16}} \\
Ours (w/ \cite{wei2024d})         & {\srate{16}{16}} & {\srate{15}{16}} & {\srate{13}{15}} & {\srate{13}{16}} \\
Ours (w/o \cite{wei2024d})          & {\srate{10}{16}}  & {\srate{6}{10}} & {\srate{3}{6}} & {\srate{3}{16}} \\
\bottomrule
\end{tabular}
\label{tab:compare_real}
\end{table}

Imitation learning baselines such as ACT exhibit a severe sim-to-real gap and rarely succeed in the real world, even in \textsc{Drag} and \textsc{Fling}. In contrast, our keypoint-conditioned primitive policy transfers zero-shot robustly. We further ablate the D3RoMA~\cite{wei2024d} depth refinement, success rates drop due to noisy single-view depth and D3RoMA~\cite{wei2024d} markedly alleviates these artifacts and restores performance by stabilizing keypoint predictions and grasp selection.

\subsection{Generalization of Garments in Simulation}

We assess material-level generalization entirely in simulation using an unseen garment mesh. The density is fixed at $1000\,\mathrm{kg}/\mathrm{m}^3$ and the Young’s modulus ranges from $300$ to $3000$ kPa, which lies within the training randomization range. We vary $E$ as a representative stiffness parameter; other randomized material/contact parameters are fixed here for clarity. As reported in Table~\ref{tab:ablate_sim}, the trained policy maintains a consistently high success rate across this stiffness range, indicating that (i) our domain-randomized data generation adequately covers the task’s material variability and (ii) the simulator serves as a reliable testbed for collecting and evaluating policies under diverse physical parameters.

\begin{table}[t]
\centering
\caption{Simulation performance across materials.}
\setlength{\aboverulesep}{0pt}
\setlength{\belowrulesep}{0pt}
\setlength{\abovetopsep}{0pt}
\setlength{\belowbottomsep}{0pt}
\setlength{\cmidrulesep}{0pt}
\begin{tabular}{r|cccc}
\toprule
\textbf{$E$ (kPa)} & $S_1$(\%)~$\uparrow$ & $S_2$(\%)~$\uparrow$ & $S_3$(\%)~$\uparrow$ & $S$(\%)~$\uparrow$ \\
\midrule
$300$      & {\srate{15}{16}}  & {\srate{14}{15}} & {\srate{12}{14}} & {\srate{12}{16}} \\
$1000$         & {\srate{16}{16}}  & {\srate{15}{16}} & {\srate{14}{15}} & {\srate{14}{16}} \\
$3000$          & {\srate{16}{16}}  & {\srate{16}{16}} & {\srate{14}{16}} & {\srate{14}{16}} \\
\bottomrule
\end{tabular}
\vspace{-5mm}
\label{tab:ablate_sim}
\end{table}

\subsection{Generalization of Garments in Real-World}

We evaluate zero-shot transfer on ten upper garments spanning variations in color, size, fabric, and collar/hem geometry
    (Fig.~\ref{fig:results}). The set includes eight sleeveless tops (\#1-\#8) and two dresses (\#9 and \#10) to stress test category shift (Fig.~\ref{fig:clothes_headshots}). Across items, mass lies in $[0.055,\,0.105]$\,kg, body length in $[57,\,70]$\,cm, hem width in $[47,\,56]$\,cm, collar-part length in $[20,\,25]$\,cm, and collar-part width in $[4,\,10]$\,cm.

We summarize results in Table \ref{table:generalization}. Overall, the policy attains a high success rate on sleeveless tops, proving that the keypoint-conditioned primitives generalize across textures and geometry changes. Success tends to increase with garment size: larger hems provide a wider access opening, which eases gripper insertion and stabilizes \textsc{Insert\&Pull}. The fabric of \#1 and \#2 is Modal\textsuperscript{\textregistered}, where pronounced wrinkling during manipulation reveals a small sim-to-real gap. For stress tests, \#9 exhibits the lowest performance because its opening is comparable to the manipulator’s effective cross-section, making insertion kinematically challenging; in contrast, \#10 demonstrates that the method reasonably generalizes to novel categories as well when geometry permits.

\textbf{Failure modes.} Rare failures arise when rollouts drift outside the primitives’ expected state distribution. The most common failure is an undersized hem opening: minor pose errors and friction during \textsc{Drag}/\textsc{Fling} or suboptimal \textsc{Insert\&Pull} keypoint selection yield an aperture that exists but is too narrow or unstable for insertion. A distinct failure is that \textsc{Drag}/\textsc{Fling} do not expose an upper-side single-layer hem region, so no opening can be formed and \textsc{Insert\&Pull} cannot be initiated. We also observe occasional perception U-Net misprediction under severe self-occlusion. A lightweight state-monitoring and recovery module, e.g., an auxiliary U-Net that classifies the current state and triggers autonomously re-canonicalization could improve robustness.

\section{Conclusion}

We presented \emph{Right-Side-Out}, a zero-shot sim-to-real framework for garment reversal that factors the task into \textsc{Drag}, \textsc{Fling}, and \textsc{Insert\&Pull}. The policy operates with depth-driven observations and compact primitives parameterization, and is supervised entirely in a high-fidelity, GPU-parallel MPM simulator that models thin-shell deformation, self-collision, and frictional robot–cloth contact. On a dual-arm platform, the approach achieves robust right-side-out performance across diverse garments, and ablations confirm the importance of task decomposition and simulator fidelity.

While we have focused on sleeveless upper garments, extending the framework to broader categories (e.g., long-sleeve shirts and pants) that introduce new topology and self-occlusion patterns is a promising next step.

\begin{table}[t] \centering
\caption{Real-world performance across garments.}
\setlength{\aboverulesep}{0pt}
 \setlength{\belowrulesep}{0pt}
 \setlength{\abovetopsep}{0pt}
 \setlength{\belowbottomsep}{0pt}
 \setlength{\cmidrulesep}{0pt}
\begin{adjustbox}{max width=\linewidth}
\setlength{\tabcolsep}{1pt}
\setlength{\extrarowheight}{0pt}
\renewcommand{\arraystretch}{0.95}
\begin{tabular}{r|*{10}{c}}
\toprule
 & \multicolumn{10}{c}{\textbf{Garment ID}} \\
\textbf{Metrics~} & \textbf{\#1} & \textbf{\#2} & \textbf{\#3} & \textbf{\#4} & \textbf{\#5} &
\textbf{\#6} & \textbf{\#7} & \textbf{\#8} & \textbf{\#9} & \textbf{\#10} \\
\midrule
$S_1$(\%)$\uparrow$ & {\srate{16}{16}} & {\srate{16}{16}} & {\srate{16}{16}} & {\srate{16}{16}} &
                   {\srate{16}{16}} & {\srate{16}{16}} & {\srate{16}{16}} & {\srate{16}{16}} &
                   {\srate{12}{16}} & {\srate{14}{16}} \\
$S_2$(\%)$\uparrow$ & {\srate{15}{16}} & {\srate{13}{16}} & {\srate{15}{16}} & {\srate{16}{16}} &
                   {\srate{15}{16}} & {\srate{14}{16}} & {\srate{13}{16}} & {\srate{14}{16}} &
                   {\srate{7}{12}}  & {\srate{10}{14}} \\
$S_3$(\%)$\uparrow$ & {\srate{12}{15}} & {\srate{11}{13}} & {\srate{13}{15}} & {\srate{16}{16}} &
                   {\srate{13}{15}} & {\srate{12}{14}} & {\srate{11}{13}} & {\srate{10}{14}} &
                   {\srate{4}{7}}   & {\srate{7}{10}}  \\
$S$(\%)$\uparrow$  & {\srate{12}{16}} & {\srate{11}{16}} & {\srate{13}{16}} & {\srate{16}{16}} &
                   {\srate{13}{16}} & {\srate{12}{16}} & {\srate{11}{16}} & {\srate{10}{16}} &
                   {\srate{4}{16}}  & {\srate{7}{16}}  \\
\bottomrule
\end{tabular}
\end{adjustbox}
\vspace{-5mm}
\label{table:generalization}
\end{table}

\small
\bibliographystyle{IEEEtran}
\bibliography{references}

\end{document}